\documentclass[aos,preprint]{imsart}

\RequirePackage{amsthm,amsmath,amsfonts,amssymb}
\RequirePackage[numbers,sort&compress]{natbib}
\RequirePackage[colorlinks,citecolor=blue,urlcolor=blue]{hyperref}%% uncomment this for coloring bibliography citations and linked URLs
\RequirePackage{graphicx}%% uncomment this for including figures
\RequirePackage[OT1]{fontenc}

\usepackage{mlmodern}
\usepackage{latexsym}
\usepackage{mathrsfs}
\usepackage{amssymb}
\usepackage{hyperref}
\usepackage{xcolor}
\usepackage{caption, subcaption}
\usepackage{stmaryrd}
\usepackage{dsfont}
\usepackage{ifthen}
\usepackage{enumerate}
\usepackage{bbm}

\usepackage{graphicx}
\usepackage{url} % not crucial - just used below for the URL 
\RequirePackage[OT1]{fontenc}
\usepackage[numbers]{natbib}
\usepackage{amsthm,amsmath}
\usepackage{graphicx,ifthen}

\startlocaldefs
\newtheoremstyle{example}{\topsep}{\topsep}%
     {}%         Body font
     {}%         Indent amount (empty = no indent, \parindent = para indent)
     {\rmfamily}% Thm head font
     {}%        Punctuation after thm head
     {\newline}%     Space after thm head (\newline = linebreak)
     {\thmname{#1}\thmnumber{ #2}\thmnote{ #3}}%         Thm head spec

   \theoremstyle{example}

\numberwithin{equation}{section}
\theoremstyle{plain}

\newtheorem{rem}{Remark}[section]

\newcommand{\Lower}[2]{\smash{\lower #1 \hbox{#2}}}
\newcommand{\ben}{\begin{enumerate}}
\newcommand{\een}{\end{enumerate}}
\newcommand{\bi}{\begin{itemize}}
\newcommand{\ei}{\end{itemize}}

\endlocaldefs

\begin{document}

\begin{frontmatter}
%%%%%%%%%%%%%%%%%%%%%%%%%%%%%%%%%%%%%%%%%%%%%%
%%                                          %%
%% Enter the title of your article here     %%
%%                                          %%
%%%%%%%%%%%%%%%%%%%%%%%%%%%%%%%%%%%%%%%%%%%%%%
\title{Learning from Neighbors with PHIBP: Predicting Infectious Disease Dynamics in Data-Sparse Environments}
%\title{A sample article title with some additional note\thanksref{T1}}
\runtitle{PHIBP}
%\thankstext{T1}{A sample of additional note to the title.}

\begin{aug}
%%%%%%%%%%%%%%%%%%%%%%%%%%%%%%%%%%%%%%%%%%%%%%%
%% Only one address is permitted per author. %%
%% Only division, organization and e-mail is %%
%% included in the address.                  %%
%% Additional information can be included in %%
%% the Acknowledgments section if necessary. %%
%% ORCID can be inserted by command:         %%
%% \orcid{0000-0000-0000-0000}               %%
%%%%%%%%%%%%%%%%%%%%%%%%%%%%%%%%%%%%%%%%%%%%%%%
\author[A]{\fnms{Edwin}~\snm{Fong}\ead[label=e1]{chefong@hku.hk}},
\author[B]{\fnms{Lancelot}~\snm{F. James}\ead[label=e2]{lancelot@ust.hk}},
\and
\author[C]{\fnms{Juho}~\snm{Lee}\ead[label=e3]{juholee@kaist.ac.kr}}
%%%%%%%%%%%%%%%%%%%%%%%%%%%%%%%%%%%%%%%%%%%%%%
%% Addresses                                %%
%%%%%%%%%%%%%%%%%%%%%%%%%%%%%%%%%%%%%%%%%%%%%%
\address[A]{Department of Statistics and Actuarial Science, HKU\printead[presep={,\ }]{e1}}
\address[B]{Department of ISOM, HKUST\printead[presep={,\ }]{e2}}
\address[C]{The Graduate School of AI, KAIST\printead[presep={,\ }]{e3}}
\end{aug}

\begin{abstract}
Modeling sparse count data, which arise across numerous scientific fields, presents significant statistical challenges. This chapter addresses these challenges in the context of infectious disease prediction, with a focus on predicting outbreaks in geographic regions that have historically reported zero cases. To this end, we present the detailed computational framework and experimental application of the Poisson Hierarchical Indian Buffet Process (PHIBP) \cite{hibp25}, a Bayesian machine learning model with demonstrated success in handling sparse count data in microbiome and ecological studies. The PHIBP’s architecture, grounded in the concept of absolute abundance, systematically borrows statistical strength from related regions and circumvents the known sensitivities of relative-rate methods to zero counts. Through a series of experiments on infectious disease data, we show that this principled approach provides a robust foundation for generating coherent predictive distributions and for the effective use of comparative measures such as alpha and beta diversity. The chapter’s emphasis on algorithmic implementation and experimental results confirms that this unified framework delivers both accurate outbreak predictions and meaningful epidemiological insights in data-sparse environments.
\end{abstract}



\end{frontmatter}

\section{Introduction}

Predicting disease prevalence in regions that have not yet reported cases—despite clear outbreaks in neighboring areas—is a problem most people now recognize from lived experience: when will it reach us, and how severe might it be? Turning this intuitive concern into reliable forecasts is a statistically challenging task. Historically, one can relate this to classes of count-based models for both realized abundance (counts) and prediction of unseen numbers and types of species in the classic work of~\cite{Fisher}. There Fisher, in answer to Corbett's question about the expected number of distinct species of butterflies to be seen in a future sample, developed the logarithmic series distribution to describe the prevalence of individually observed species. This then translates into the remarkable work of~\cite{JMQ}, which empirically verifies the Negative Binomial distribution as a viable model for soil microorganism counts in the form of a mixed Poisson variable: the framework in~\cite{JMQ} provides an empirical justification for a model in which the number of colonies is distributed according to a Poisson distribution, while the bacterial counts within each colony independently follow the logarithmic series distribution introduced by~\cite{Fisher}.

Naturally, while many core elements of those works translate to modern datasets, they do not capture the notion of sharing information across groups, nor matters of count sparsity. In particular, while the prevalence of zeros is common in modern datasets, we further highlight the notion of sparse co-occurrence, where samples across groups do not share many common species. These challenges, along with modeling complex multivariate count distributions, are all addressed by the Poisson Hierarchical Indian Buffet Process (PHIBP)~\cite{hibp25} as applied to complex microbiome sampling models. Here, we focus on its ability to pool information across groups to handle zero observations, leading to credible predictive and inferential modeling for infectious disease prediction. 

The present work is part of a three-paper arc. The foundational work~\cite{hibp25} establishes PHIBP as a comprehensive statistical framework for hierarchical count data. Its theoretical complement~\cite{james2025coagfragduality} resolves long-standing problems in coagulation–fragmentation duality, revealing deep structural properties of the framework that extend to continuous-time processes and complex ancestral constructions. This application demonstrates PHIBP's practical capabilities for disease surveillance while simultaneously showcasing the broader extensibility of the trilogy: the same underlying architecture accommodates both the esoteric theoretical developments and domain-specific customizations amenable to future AI/LLM frameworks for intelligent model specification.

Before detailing the model, we first outline several architectural innovations that distinguish PHIBP and motivate its use for disease prediction. As the theory of PHIBP has been detailed previously, our focus is on methodology and computational inference within the present context of disease prediction.

\section{Architectural Innovations of PHIBP}

\subsection{Absolute abundance vs.\ compositional rates}

A fundamental design choice in count-based modeling is whether to work with absolute abundances (raw counts or rates) or relative abundances (proportions summing to unity within each sample). While compositional approaches—dominant in modern microbiome and ecological analyses—offer certain computational conveniences, they suffer from the closure problem: because relative abundances must sum to one, all components are mechanically constrained such that an increase in one component necessitates a proportional decrease in others, even when their absolute counts remain unchanged. This induces spurious negative correlations among components that reflect the mathematical constraint rather than genuine biological or epidemiological relationships. For infectious disease surveillance, where distinguishing true co-occurrence patterns from measurement artifacts is critical, this confounding is particularly problematic.

Consider a concrete example: two counties each report 100 total disease cases in Year~1, with County~A showing 60 cases of Disease~X and 40 of Disease~Y, while County~B shows 50 cases each. In Year~2, both counties experience identical 20-case increases in Disease~X (reaching 80 and 70 cases respectively), while Disease~Y remains stable at 40 and 50 cases. The absolute data reveal a shared trend: Disease~X is spreading identically in both locations. Yet compositional methods see County~A shifting from $60\%$/$40\%$ to $67\%$/$33\%$ and County~B from $50\%$/$50\%$ to $58\%$/$42\%$—artificially suggesting that Disease~Y is declining when its absolute burden is stable. When data contain many zeros, the distortions compound: a single new case drastically alters all proportions, and counties with zero cases cannot be meaningfully compared under relative-abundance frameworks.

The absolute-abundance perspective has deep roots in classical statistical ecology. Fisher, Corbet, and Williams (1943) modeled butterfly species abundances using the logarithmic series distribution for the frequency of species with $1,2,3,\dots$ individuals in a fixed sample, estimating the underlying abundance parameter from the observed counts rather than working with normalized compositional proportions. This foundational work then translates into the remarkable empirical investigations of~\cite{JMQ}, which verified the negative binomial as a viable model for soil microorganism counts in the form of a mixed Poisson variable. The framework in~\cite{JMQ} provides an empirical justification for a model in which the number of colonies is distributed according to a Poisson distribution, while the bacterial counts within each colony independently follow the logarithmic series distribution introduced by Fisher. Both approaches model absolute counts and treat each species' abundance as an independent realization, avoiding the sum-to-one constraint. This classical tradition recognizes that the proper modeling target is the intensity (mean count per unit exposure), not the proportion of a fixed total. The PHIBP foundations are strongly connected to, and influenced by, Pitman's manuscript~\cite{PitmanPoissonMix}, which revisits this Fisher–McCloskey perspective and highlights it as a neglected strand in the species sampling literature.
 
The PHIBP framework inherits and extends this absolute-abundance paradigm to hierarchical settings. At the global level, each disease type $\ell$ is assigned an absolute rate $\lambda_\ell > 0$ representing its mean intensity. Locally, county~$j$ experiences a transformed rate $\sigma_{j,\ell}(\lambda_\ell)$ obtained via a subordinator process that modulates the global rate according to region-specific characteristics captured by the Lévy measure~$\tau_j$. Crucially, these rates are not constrained to sum to any fixed value: a high rate for Disease~A in County~1 does not mechanically reduce the inferred rate for Disease~B, nor does it affect Disease~A's rate in County~2. The subordinator architecture preserves marginal independence of disease intensities~\cite{hibp25}.

This absolute-abundance framework has immediate implications for interpreting zero counts. Under compositional approaches, a county reporting zero cases of Disease~A and five of Disease~B is forced into the representation~(0\%, 100\%), which is undefined when computing log-ratios and creates computational difficulties. More fundamentally, compositional methods cannot distinguish sampling zeros—where Disease~A is present at low rate but unobserved—from structural zeros, where it genuinely cannot occur. PHIBP's absolute-rate parameterization accommodates both: a county can have small but nonzero~$\sigma_{j,\ell}(\lambda_\ell)$ yielding zero observed counts via Poisson variability, or~$\sigma_{j,\ell}(\lambda_\ell) \approx 0$ indicating true absence. The posterior distribution over these rates, informed by hierarchical borrowing, assigns appropriate uncertainty and enables probabilistic prediction of emergence in currently zero-count locations.

By returning to Fisher's absolute-abundance tradition and extending it through subordinator-driven hierarchical rates, PHIBP aligns the mathematical structure with the epidemiological question: not ``what proportion of cases are Disease~X?'' but ``how many cases of Disease~X do we expect per unit population?''

\subsection{Thinning vs.\ exchangeable sampling}

A second architectural choice is whether to generate observations by thinning a Poisson random measure or by exchangeable sampling from a normalized random measure. In classical exchangeable partition models, one first constructs a random probability measure $G$ and then obtains data by drawing a fixed sample of size $n$ i.i.d.\ from $G$. This induces elegant partition laws, but it implicitly treats $n$ as the fundamental quantity: the model is conditioned on the total number of individuals or reads observed. For sequencing-based surveillance or routine reporting systems, this conditioning is artificial. The total count is itself a random outcome of the measurement process, driven by laboratory throughput, reporting practices, or population size, rather than a fixed design parameter.

The PHIBP framework instead follows the thinning perspective inherent in its mixed Poisson construction. Starting from global mean intensities on the disease types, it generates local group-level intensities, and only then produces observed counts by Poisson sampling, with rates scaled by the relevant exposure (e.g.\ population-at-risk, test volume, or reporting effort). In this view, the primary random objects are the underlying intensities and the exposure; the total count is a derived random quantity. This aligns more naturally with infectious disease settings, where one does not control the number of cases, but rather observes whatever the system produces over a given period.

This distinction has several practical consequences. First, thinning provides a transparent representation of sampling zeros versus structural zeros: a disease may have strictly positive but small intensity in a county and still yield zero observed cases by chance, whereas an intensity that is shrunk toward zero across the hierarchy corresponds to genuine absence. Exchangeable sampling on normalized measures, by contrast, operates on proportions and a fixed $n$, and has no native mechanism to represent the random occurrence of zeros at fixed exposure. Second, thinning leads directly to additive prediction rules. Adding a new surveillance period or a new region corresponds to drawing new Poisson counts at the same latent intensities; the three-way decomposition into (i) completely new disease types, (ii) newly observed manifestations of existing types, and (iii) additional counts for already-observed types follows from standard Poisson calculus, as in the PHIBP prediction rules.

Finally, thinning is what makes the PHIBP-based architecture computationally usable in our setting: the mixed Poisson and compound Poisson representations depend critically on viewing the data as Poisson thinnings of underlying rate processes, rather than as exchangeable samples of fixed size. This is precisely what allows us to derive explicit posterior and predictive laws for sparse, hierarchically structured counts.

Taken together, these choices place PHIBP firmly in the absolute-abundance tradition of Fisher, Corbet, and Quenouille, where intensities and mixed Poisson structure are the primary modelling objects, while at the same time marking a genuine paradigm shift. The framework extends that tradition to hierarchical, multi-group settings with explicit rate coupling and exact joint laws for complex dependence structures—features that are not available in classical relative-abundance or purely exchangeable formulations.

\begin{remark}
For a deeper look at thinning-based sampling mechanisms as primary, exchangeability as a subsequent design choice, and the flexibility to design hierarchical models on abstract Polish spaces using $h$-biased measures with domain-specific interpretations, readers are invited to consult~\cite[Section~7]{james2025coagfragduality}.
\end{remark}

\subsection{Extensibility of the PHIBP framework}

Although not emphasized as a central structural feature, the PHIBP framework subsumes the capabilities of hierarchical compositional models such as the Hierarchical Dirichlet Process (HDP)~\cite{HDP} and its extensions. In other words, PHIBP can accomplish everything these classical models can do—but the converse is not true. This generality is made strikingly apparent in the coagulation–fragmentation duality work~\cite[see e.g.\ Sections~1.3, 3, and Proposition~2.5]{james2025coagfragduality}.

The present exposition should be viewed as part of a three-paper arc. The foundational work~\cite{hibp25} establishes PHIBP as a comprehensive statistical framework for species sampling models and hierarchical count data. Its theoretical complement~\cite{james2025coagfragduality} resolves long-standing problems in coagulation–fragmentation duality for both static and continuous-time settings, operating primarily within probability theory while revealing substantial capabilities for complex applications. The current disease surveillance application demonstrates aspects of this framework in action, though it operates within the essential skeleton of PHIBP itself. Our intent—as in the foundational work—is to provide a principled building block on which domain experts can construct bespoke extensions tailored to their specific problems.

Section~7 of~\cite{james2025coagfragduality} is particularly relevant here: it describes how to design customized PHIBP variants by directly specifying the behavior of Poisson random measure components in ways that carry concrete meaning for epidemiologists, ecologists, or other domain practitioners—rather than selecting a modeling component, such as a Gamma process or stable subordinator, merely because it yields a tractable closed-form distribution. As developed in that work, PHIBP variants now manifest as a $4 \times J$ component dual system—simultaneously encoding fine and coarse partitions, and fragmentation and coagulation operators—that naturally elevates to continuous-time models and supports sophisticated ancestral constructions such as structured Ancestral Recombination Graph (ARG) and Time to the Most Recent Common Ancestor (TMRCA) frameworks. Within this cloud-duality setup, domain experts may encode substantive structure directly through the choice of state spaces \(S_j\), mean measures, and \(h_j\)-biasing functions on point clouds, yielding models whose components are interpretable in scientific terms rather than imposed for distributional convenience. With this architecture, modeling choices can be driven by interpretability and domain insight rather than mathematical tractability alone, and the framework is not bound to standard sampling schemes such as exchangeability~\cite{hibp25}. Notably, these capabilities align naturally with the evolving potential of AI/LLM systems for model specification, allowing domain experts to translate conceptual knowledge—for example, about ARG topologies, recombination counts, or TMRCA behavior—into explicit prior structure within a principled probabilistic framework.

\section{The PHIBP model}

\textbf{Modeling objectives and contribution.} This chapter applies the PHIBP framework to infectious disease prediction, demonstrating its computational implementation in a setting where many county-disease pairs exhibit zero observed counts. The primary inferential targets are the latent disease-specific rates—global rates $H_\ell$ and local rates $\sigma_{j,\ell}(\lambda_\ell)$ for each region $j$ and disease type $\ell$—which govern the intensity of disease occurrence. From these rates, we derive posterior distributions for alpha- and beta-diversity measures and construct predictive distributions for diseases not yet observed in specific counties. The hierarchical structure enables information borrowing: counties with zero cases of a disease can still receive informative posteriors by pooling evidence from neighboring regions and the global disease catalog. While the foundational theory is developed in~\cite{hibp25}, the present application demonstrates that the PHIBP framework operates without modification on epidemiological data that appear structurally distinct from the microbiome context, revealing the framework's generality and flexibility. Furthermore, the architecture admits substantial extensions—hierarchical geographic structures, dynamic surveillance, and integration with domain-specific covariates—that we outline in Section~\ref{sec:extensions}, positioning PHIBP as a versatile building block for hierarchical count modeling across scientific domains.

We introduce the PHIBP framework for modeling hierarchical count data. For a comprehensive treatment of the underlying technical foundations—including Poisson random measures, completely random measures, and L\'evy subordinators—the reader may consult standard references such as \cite{Bertoin1999, Sato2013}. Furthermore, while we aim for a self-contained exposition, the reader is encouraged to consult the more comprehensive treatment in the primary PHIBP reference~\cite{hibp25}.

Before detailing the hierarchical structure, we briefly recall the foundational object. A \emph{Poisson random measure} (PRM) $N$ over a Polish space $\Omega$ is a random measure with the following properties:
\begin{enumerate}
\item[(i)] For disjoint measurable sets $A,B \subset \Omega$, the random variables $N(A)$ and $N(B)$ are independent.
\item[(ii)] For any measurable set $A$, $N(A) \sim \mathrm{Poisson}(\nu(A))$, where $\nu$ is a $\sigma$-finite measure on $\Omega$ called the \emph{mean measure}, satisfying $\mathbb{E}[N(A)] = \nu(A)$.
\end{enumerate}

While random measures may seem daunting, the key insight is that they can be thought of in terms of random variables or random vectors: $N(A)$ is simply a Poisson random variable whose intensity depends on the set $A$. This perspective makes them interpretable and easily incorporated into various applications. This same interpretability carries over to functionals of PRMs, such as completely random measures (CRMs), which we describe next. Intuitively, a CRM is a random intensity measure constructed by placing random jumps at random locations on a Polish space, with the jumps governed by a L\'evy measure. The marginal distributions of such measures are based on familiar infinitely divisible random variables such as Gamma, stable, and generalized Gamma, as well as more exotic choices. A natural way of characterizing such variables is via Laplace transforms, and this together with independence over disjoint sets is enough to characterize these processes.

The ability to operate on abstract Polish spaces is not only for theoretical flexibility but enables bespoke designs for applications by domain experts, as illustrated in Section~7 of~\cite{james2025coagfragduality}. Here, we use $\Omega$ generically throughout. For disease applications, $\Omega$ is a finite or countable catalogue of circulating disease types (pathogens, strains, etc.) in the population; more abstractly, it indexes data labels or types that inherit structure from the hierarchical specification.

A \emph{completely random measure} (CRM) is constructed from a PRM on the product space \((0,\infty)\times\Omega\) by integrating out the jump component. Specifically, let \(N\) be a PRM on \((0,\infty)\times\Omega\) with mean measure \(\nu(ds,dy) = \tau(ds)\,F_0(dy)\), where \(\tau\) is a L\'evy measure on \((0,\infty)\) satisfying \(\int_{(0,\infty)} \min(s,1)\,\tau(ds) < \infty\), and \(F_0\) is a \(\sigma\)-finite measure on \(\Omega\). Throughout, we write L\'evy measures in density form, for example \(\tau(ds) = \tau(s)\,ds\), and adopt this compact notation for convenience.

Then the CRM is the random atomic measure
\[
B_0 \;=\; \int_{(0,\infty) \times \Omega} \lambda \, N(d\lambda, dy) \;=\; \sum_{l \ge 1} \lambda_l \,\delta_{Y_l},
\]
where the sum ranges over the (almost-surely countable) atoms $(\lambda_l, Y_l)$ of $N$, and we write $B_0 \sim \mathrm{CRM}(\tau, F_0)$. The L\'evy measure $\tau$ governs the distribution of jump sizes $(\lambda_\ell)$, and its Laplace exponent is $\psi(\gamma) := \int_{(0,\infty)} (1 - e^{-\gamma s})\,\tau(ds)$ for $\gamma \ge 0$. 

For the PHIBP framework developed here, we choose $F_0$ to be a probability measure on $\Omega$—this means we do not need to track a normalizing constant in the mean intensity.

We consider $J$ related regions (e.g.\ counties), $j\in[J]$, each with $M_j$ replicated count samples (e.g.\ years), $i\in[M_j]$. A sample has the form
\[
Z^{(i)}_j \;=\; \sum_{l\ge1} N^{(i)}_{j,l}\,\delta_{Y_l},
\]
where $Y_l$ indexes a global catalogue of disease types and $N^{(i)}_{j,l}\in\{0,1,2,\dots\}$ is the observed count of disease type $Y_l$ in sample $i$ from region $j$. The PHIBP specifies a hierarchical prior on the latent mean abundance rates that generate these counts. At the top level, we draw a global completely random measure
\[
B_0 \;=\; \sum_{l\ge1} \lambda_l \,\delta_{Y_l} \;\sim\; \mathrm{CRM}(\tau_0,F_0),
\]
where $\tau_0$ is a L\'evy measure on $(0,\infty)$ and $F_0$ a non-atomic probability measure on the Polish space $\Omega$ of disease types. The jumps $(\lambda_l)$ are interpreted as global mean rates of the diseases $(Y_l)$ across all regions.

Given $B_0$, each region $j$ has a local CRM
\[
B_j \mid B_0 \;\sim\; \mathrm{CRM}(\tau_j,B_0), \qquad j\in[J],
\]
with L\'evy measure $\tau_j$ on $(0,\infty)$. When $B_0$ is not a probability measure, this is well-defined as follows: we condition on the realized atoms $(\lambda_\ell, Y_\ell)$ of $B_0$ and construct the local jumps accordingly. Conditionally on $B_0=\sum_l \lambda_l\delta_{Y_l}$ we may write
\[
B_j \;\stackrel{d}{=}\; \sum_{l\ge1} \sigma_{j,l}(\lambda_l)\,\delta_{Y_l}, \qquad \sigma_{j,l}(\lambda_l)\stackrel{d}{=}\sigma_j(\lambda_l),
\]
where $\sigma_{j,l}(\lambda_l) = \sum_{k \ge 1} s_{j,k,l}$ and the atoms $(s_{j,k,l})_{k \ge 1}$ are the jumps of the subordinator $\sigma_j$ restricted to the interval of length $\lambda_l$. More formally, $(\sigma_j(t):t\ge 0)$ is a subordinator with L\'evy measure $\tau_j$, satisfying for $s<t$, $\sigma_{j}(t)-\sigma_{j}(s)\overset{d}{=}\sigma_{j}(t-s)$ independent of $\sigma_{j}(s)$. Thus $\lambda_l$ is the global rate of disease type $Y_l$, and $\sigma_{j,l}(\lambda_l) = \sum_{k \ge 1} s_{j,k,l}$ is its local mean rate in region $j$, decomposed into finer-grained subspecies-level intensities $(s_{j,k,l})$. This construction—using $B_0$ as the directing measure for $B_j$—is what allows the local rates to inherit the disease labels $(Y_l)$ from the global level while modulating their intensities via the subordinator, with the additional structure that each disease type $Y_l$ harbors internal heterogeneity captured by the subspecies markers indexed by $k$.

Throughout, as in~\cite{hibp25}, we denote by $\psi_j(\cdot)$ and $\Psi_0(\cdot)$ the Laplace exponents~\cite{Bertoin1999} associated with the L\'evy measures $\tau_j$ and $\tau_0$, respectively. For a subordinator $\sigma_j$ with L\'evy measure $\tau_j$, the Laplace exponent $\psi_j$ is defined by

\[
\mathbb{E}[e^{-\gamma_{j}\sigma_{j}(t)}]=e^{-t\psi_{j}(\gamma_{j})},
\]
where
\[
\psi_j(\gamma_{j})
\;=\;
\int_{(0,\infty)} \big(1-e^{-\gamma_{j} s}\big)\,\tau_j(ds).
\]
Similarly, for the global subordinator $\sigma_0$ with L\'evy measure $\tau_0$,
\[
\Psi_0(\gamma_{0})
\;=\;
\int_{(0,\infty)} \big(1-e^{-\gamma_{0}\lambda}\big)\,\tau_0(d\lambda),
\quad \gamma_{0}\ge 0.
\]
The Laplace exponent $\psi_j(\gamma_j)$ depends only on $\gamma_j$ and the L\'evy measure $\tau_j$, not on any particular time $t$ or magnitude $\lambda$. We assume that $\int_{0}^{\infty}\min(s,1)\tau_{j}(ds)<\infty$ ensuring the finiteness of $B_{j}(\Omega)$ for $j=0,\ldots,J$. In practice, we evaluate $\psi_j$ at aggregated exposures such as $\psi_j\!\big(\sum_{i=1}^{M_j}\gamma_{i,j}\big)$ and compose exponents as $\Psi_0\big(\sum_{j=1}^J\psi_j(\cdot)\big)$, exactly as in the PHIBP construction of~\cite{hibp25}. 

Furthermore, setting
\[
\psi^{(c)}_{j}(\gamma_{j}) = \int_{0}^{\infty} s^{c} e^{-\gamma_{j}s}\,\tau_{j}(ds),
\]
we have the mixed truncated Poisson distributions with law denoted by
\(\mathrm{MtP}(\tau_{j},\gamma_{j})\) and probability mass function, for
\(c=1,2,\ldots\), given by
\[
\frac{\gamma^{c}_{j}\,\psi^{(c)}_{j}(\gamma_{j})}{\psi_{j}(\gamma_{j})\,c!},
\]
where
\[
\sum_{c=1}^{\infty}\frac{\gamma^{c}_{j}\,\psi^{(c)}_{j}(\gamma_{j})}{c!}
= \psi_{j}(\gamma_{j}).
\]

The PHIBP $(Z^{(i)}_j, i\in[M_{j}], j\in[J])$ is specified as a mixed Poisson point process (PoiP) driven by the local CRMs; a Poisson point process is the random counting measure generated by a CRM, counting events according to the intensities encoded in the measure. For each $j$ and $i\in[M_j]$,
\begin{equation}
Z^{(i)}_j \,\big|\, B_j \;\stackrel{\mathrm{ind}}{\sim}\; \mathrm{PoiP}(\gamma_{i,j} B_j),
\label{eq:phibp-PoiP}
\end{equation}
where $\gamma_{i,j}>0$ is a sample-specific exposure. Using the atomic form of $B_j$, this means
\begin{equation}
Z^{(i)}_j \;\stackrel{d}{=}\; \sum_{l\ge1} \mathscr{P}^{(i)}_{j,l}\big(\gamma_{i,j}\sigma_{j,l}(\lambda_l)\big)\,\delta_{Y_l},
\label{eq:mixed-Poisson-rep}
\end{equation}
with conditionally independent Poisson counts
\[
\mathscr{P}^{(i)}_{j,l}\big(\gamma_{i,j}\sigma_{j,l}(\lambda_l)\big)\;\sim\;\mathrm{Poisson}\!\big(\gamma_{i,j}\sigma_{j,l}(\lambda_l)\big).
\]
This is the basic mixed Poisson representation: each observed count $N^{(i)}_{j,l}$ is an outcome of a Poisson random variable with mean $\gamma_{i,j}\sigma_{j,l}(\lambda_l)$, the product of exposure and local intensity.

Summing over samples within region $j$ gives
\[
\sum_{i=1}^{M_j} Z^{(i)}_j \,\bigg|\, B_j \;\sim\; \mathrm{PoiP}\!\Big(\Big(\sum_{i=1}^{M_j}\gamma_{i,j}\Big) B_j\Big)
\]
and induces the usual quantity
\[
\psi_j\!\Big(\sum_{i=1}^{M_j}\gamma_{i,j}\Big)
\;=\;
\int_{(0,\infty)} \Big(1-e^{-s\sum_{i=1}^{M_j}\gamma_{i,j}}\Big)\,\tau_j(ds),
\]
which governs the rate of distinct disease types (atoms) observed in region $j$ after aggregating over its $M_j$ samples. Writing $N_{j,l} := \sum_{i=1}^{M_j} N^{(i)}_{j,l}$ for the total count of disease type $Y_l$ in region $j$, we have
\[
N_{j,l} \,\big|\, (\sigma_{j,l}(\lambda_l))_{j,l} \;\sim\; \mathrm{Poisson}\!\Big(\Big(\sum_{i=1}^{M_j}\gamma_{i,j}\Big)\sigma_{j,l}(\lambda_l)\Big),
\]
independently over $(j,l)$. The key result (Theorem~3.1 of~\cite{hibp25}) shows that the joint law of the summed processes
\begin{equation}\label{eq:joint-law}
\Big(\sum_{i=1}^{M_j} Z^{(i)}_j,\; j\in[J]\Big)
\;\overset{d}{=}\;
\Big(\sum_{\ell=1}^{\varphi} \tilde{N}_{j,\ell}\,\delta_{\tilde{Y}_{\ell}},\; j\in[J]\Big)
\;\overset{d}{:=}\;
\Big(\sum_{\ell=1}^{\varphi} \big[\sum_{k=1}^{X_{j,\ell}} C_{j,k,\ell}\big]\delta_{\tilde{Y}_{\ell}},\; j\in[J]\Big)
\end{equation}
admits an exact compound Poisson representation in terms of a random number of latent (sub)-species level clusters and their counts. Here $\varphi$ is the random number of distinct disease types that appear across all regions, and has a Poisson distribution with mean
\[
\Psi_0\big(\sum_{j=1}^J \psi_j(\sum_{i=1}^{M_j}\gamma_{i,j})\big).
\]
For each observed disease type $\ell$, there is a global posterior mean rate $H_\ell>0$. The number of operational taxonomic units (OTUs)—representing the finest-grained hierarchical classification within each disease type—in region $j$ is denoted $X_{j,\ell}$, with total across all regions $X_\ell := \sum_{j=1}^J X_{j,\ell}$. Jointly, $(H_\ell,X_\ell)$ has the mixed Poisson–MtP structure of~\cite{hibp25}, with $X_\ell \sim \mathrm{MtP}\big(\tau_0, \sum_{j=1}^J \psi_j(\sum_{i=1}^{M_j}\gamma_{i,j})\big)$ and $H_\ell \mid X_{\ell}=x_{\ell}$ conditionally distributed with density proportional to
\[
\lambda^{x_\ell}\exp\!\{-\lambda\sum_{j=1}^J\psi_j(\sum_{i=1}^{M_j}\gamma_{i,j})\}\tau_0(\lambda).
\]
Given $X_\ell=x_\ell$, the allocation of OTU clusters across regions is multinomial,
\[
(X_{1,\ell},\dots,X_{J,\ell}) \mid X_\ell=x_\ell \sim \mathrm{Multinomial}(x_\ell; q_1,\dots,q_J),
\]
with $q_j \propto \psi_j\big(\sum_{i=1}^{M_j}\gamma_{i,j}\big)$. Finally, given these allocations and the local Lévy measures $\tau_j$, the OTU-level cluster sizes $C_{j,k,\ell}$ are independent mixed truncated Poisson variables
\[
C_{j,k,\ell} \sim \mathrm{MtP}\big(\tau_j,\sum_{i=1}^{M_j}\gamma_{i,j}\big),
\]
so that each regional total $\tilde{N}_{j,\ell} = \sum_{k=1}^{X_{j,\ell}} C_{j,k,\ell}$ is a sum of i.i.d.\ MtP components driven by the global rate $H_\ell$. The joint vector of random counts $(\tilde{N}_{j,\ell}, j\in[J])$ is independent over $\ell\in[\varphi]$ and is shown to have an explicit, albeit more complex, form within the MtP family in~\cite[Proposition~4.3]{hibp25}, and plays important roles in~\cite[Section~2]{james2025coagfragduality}.

\begin{rem}
The jump decomposition $(s_{j,k,\ell})$ underlying the local rates $\sigma_{j,\ell}(\lambda_\ell)$ admits representation in terms of $h$-biased measures, a framework developed in Section~7 of~\cite{james2025coagfragduality}. Specifically, one may write $s_{j,k,\ell}=h_{j}(\mathbf{e}_{j,k,\ell})$, where $(\mathbf{e}_{j,k,\ell})$ are points of a Poisson random measure on more general spaces $S_j$, and $h_j$ are non-negative functions. This flexibility allows domain experts to encode substantive structure directly—through the choice of $S_j$ and $h_j$—rather than selecting components solely for mathematical convenience. In disease surveillance, for instance, $S_j$ might index transmission pathways, and $h_j$ might encode geographic proximity or demographic factors, yielding interpretable hierarchical structures tailored to epidemiological questions.
\end{rem}

\subsection{Predicting the unseen and alpha/beta diversities}\label{sec:predict_ab}

We now recount some of the novel developments in~\cite{hibp25} that will figure prominently in our analysis. The representation in~\eqref{eq:joint-law} decomposes the pairs of rates and species labels $(\lambda_{l},Y_{l})_{l\ge 1}$ into two components: unobserved species with rates $(\lambda'_{l},Y'_{l})_{l\ge 1}$ and the $\varphi$ observed species $\tilde{Y}_{\ell}$ with posterior mean rates $H_{\ell}$ for $\ell\in[\varphi]$. We use this compound Poisson representation, together with the posterior local rates $\sigma_{j,\ell}(H_\ell)$, to define and compute Bayesian alpha- and beta-diversity measures and disease prediction functionals.

Within this framework, we define alpha-diversity (within-group diversity) via the Shannon entropy:
\begin{equation}
\mathscr{D}_{j} := -\sum_{l=1}^{\varphi} \frac{\tilde{\sigma}_{j,l}(H_{l})}{\sum_{t=1}^{\varphi} \tilde{\sigma}_{j,t}(H_{t})} \log\left(\frac{\tilde{\sigma}_{j,l}(H_{l})}{\sum_{t=1}^{\varphi} \tilde{\sigma}_{j,t}(H_{t})}\right).
\label{Shannon1}
\end{equation}
For beta-diversity (between-group diversity), we define a dissimilarity based on the Bray-Curtis index:
\begin{equation}
\mathscr{B}_{j,v} := \frac{\sum_{l=1}^{\varphi} |\tilde{\sigma}_{j,l}(H_{l})-\tilde{\sigma}_{v,l}(H_{l})|}{\sum_{l=1}^{\varphi} (\tilde{\sigma}_{j,l}(H_{l})+\tilde{\sigma}_{v,l}(H_{l}))}, \quad j\neq v \in[J].
\label{BrayCurtis1}
\end{equation}

Our constructs, while new, are in the form of more classical measures of diversity as detailed in~\cite{RICOTTA2021, willis2022estimating}. In ecological and epidemiological settings, \emph{alpha-diversity} measures the within-region richness and evenness of types—in our case, how many diseases circulate in a county and how evenly their burdens are distributed—while \emph{beta-diversity} quantifies between-region dissimilarity, capturing how disease profiles differ across counties or how transmission patterns diverge geographically. 

Classically, these metrics are computed from observed proportions, but such compositional methods break down in the presence of many zeros, causing artificial inflation of dissimilarity when low-prevalence types are simply unobserved. The PHIBP framework replaces fixed proportions with \emph{latent abundance rates}, producing alpha- and beta-diversities that properly distinguish sparse detection from true absence. Moreover, these quantities are Bayesian random variables, yielding full posterior distributions that reflect uncertainty—critical when data are sparse or uneven across regions—rather than single point estimates. 

This same decomposition underlies PHIBP's approach to the unseen-disease problem, enabling principled prediction of diseases not yet observed but likely to appear under continued sampling. The random variable $\varphi$ in~\eqref{eq:joint-law} represents the total number of disease types that could appear in the observed data, including both those actually detected and those present in the environment but unobserved due to finite sampling. The posterior distribution of the unobserved rates $(\lambda'_{l})_{l\ge 1}$ provides a basis for predicting the prevalence and regional distribution of diseases that have not yet been observed but may emerge with additional surveillance effort or in future samples.

In our previous PHIBP experiments on the Dorado Outcrop microbiome data, originally investigated by~\cite{lee2015microbial, willis2022estimating}, this Bayesian, rate-based treatment proved essential: the model captured rare shared taxa, avoided pseudocount artifacts, and produced more stable alpha- and beta-diversity estimates than compositional alternatives. The same advantages carry over directly to the disease-prediction setting studied here.

\section{Infectious disease dataset}

The dataset of focus in this chapter is the `Infectious Diseases by Disease, County, Year, Sex' dataset collected by the California Department of Public Health \cite{CHHS}. The dataset contains the prevalence of a selection of communicable infectious diseases across all 58 counties within California from 2001 - 2023 obtained through local health providers and laboratories. Due to factors such as incomplete reporting by healthcare providers or lack of access to healthcare, the reported prevalence of many diseases is likely an underestimate of the true disease prevalence, so we do not expect underlying rates to actually be zero.  A key property of the dataset is thus count sparsity, where there are multiple regions with no reported counts of specific diseases. For example, in 2023, there were 144 reported cases of Listeriosis, but these cases were all confined to occur in only 28 out of 58 counties. As the diseases are communicable, there is also the expectation of some prevalence correlations occurring due to geographical vicinity, which we will quantify through the aforementioned beta-diversity and visualize in a geographical heatmap. 

 The main goal of our analysis is to predict disease counts in each county, paying careful attention to county-disease pairs that have 0 reported counts.  We will then proceed to validate these predictions in a held-out test set to highlight the advantages of information borrowing between counties. A more exploratory goal is to evaluate whether or not geographical vicinity is a driver in correlation of the infectious disease landscape. To carry out the latter task, we will compute the estimated beta-diversities for different reference counties and visualize the values on a geographical heatmap.

\subsection{Experimental details}\label{sec:exp_det}

For our analysis, we begin by filtering out highly common infectious diseases to focus our study on rarer diseases. This includes disease such as as Salmonellosis and Campylobacteriosis which have a maximum count of over 1000 in a single county, leaving us with 36 remaining infectious diseases. We will then fit the PHIBP model to the years 2001 - 2014, treating the remaining years 2015-2023 as the held-out test dataset.  

For the PHIBP model, we specify both $\tau_0$ and $\tau_j$ as L\'{e}vy measures of a Generalized Gamma (GG) subordinator. Each measure is parameterized as
\[
\tau_j(\lambda) = \frac{\theta_j}{\Gamma(1-\alpha_j)}\lambda^{-1-\alpha_j}e^{-\lambda} \text{ for } j \in \{0\} \cup [J],
\]
where $\theta_j > 0$ and $\alpha_j \in [0,1)$ for all $j \in {0} \cup [J]$. The parameter $\alpha_j$ controls the degree of population diversity: larger values of $\alpha_j$ induce richer latent structures, resulting in a more diverse collection of disease types appearing within or across counties. A notable special case arises when $\alpha_j = 0$, in which case the GG subordinator reduces to the Gamma (GA) subordinator. Since GA subordinators exhibit distinct asymptotic behavior, we treat GA as a separate baseline model and compare it to the GG specification with $\alpha_j > 0$.

For posterior inference, we run three independent MCMC chains for both GG and GA versions of the PHIBP model. Each chain is executed for 40,000 iterations, with the first 20,000 iterations discarded as burn-in. We retain samples using a thinning interval of 10 to reduce autocorrelation. Additional implementation details—including prior specification, sampling scheme, and prediction procedures —are provided in \cite{hibp25}.

% {\color{blue} Add prior/sampler setup here?}

\subsection{Results}

\begin{figure}
    \centering
    \includegraphics[width=0.32\linewidth]{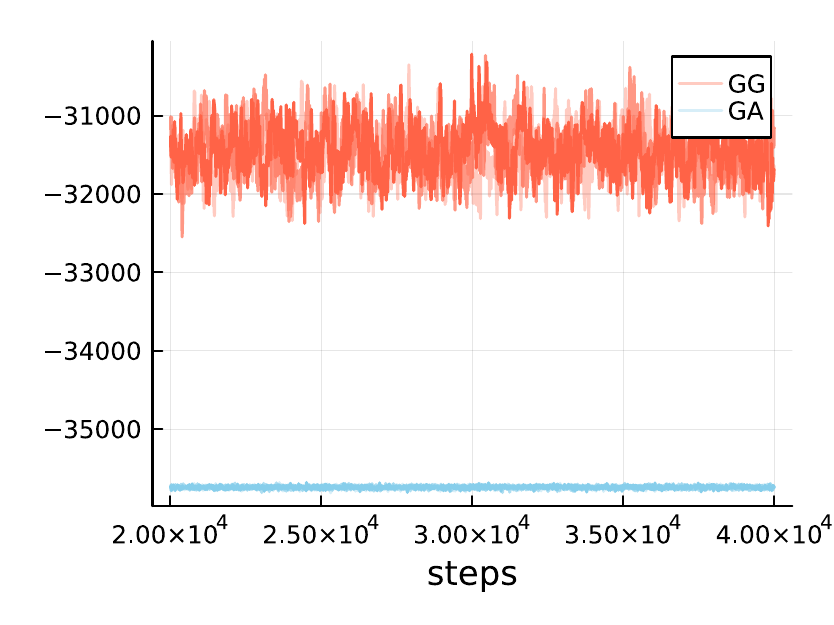}
    \includegraphics[width=0.32\linewidth]{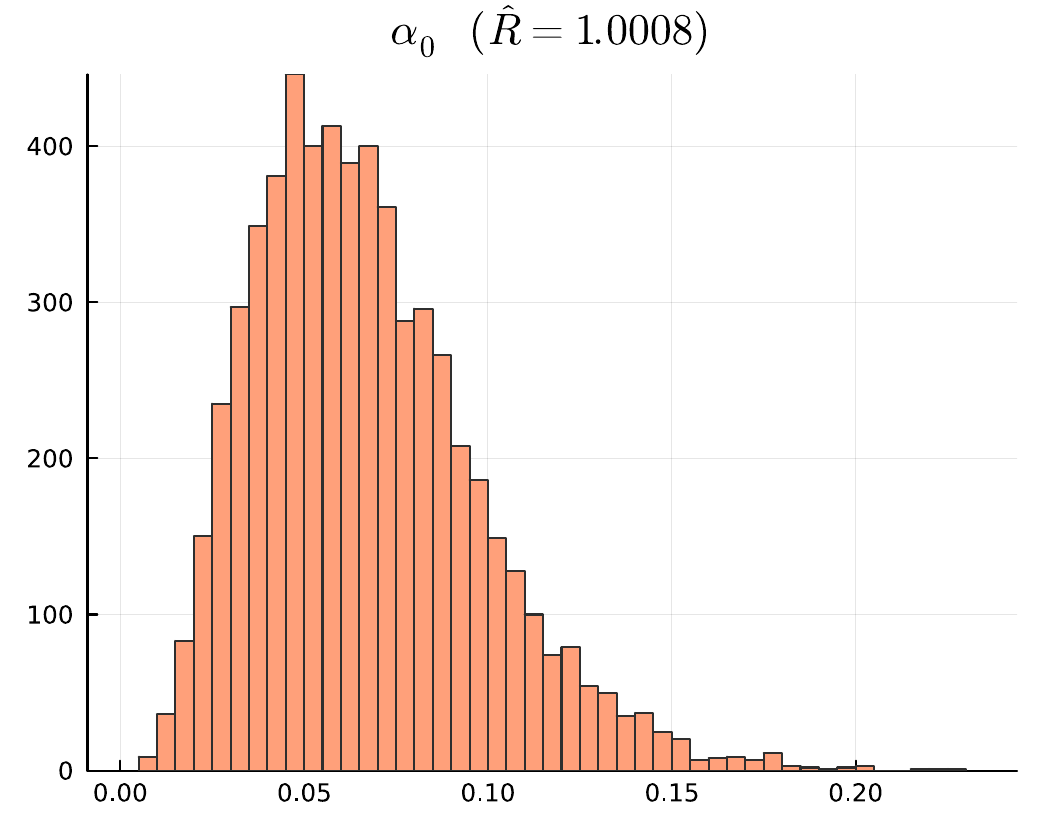}
    \includegraphics[width=0.32\linewidth]{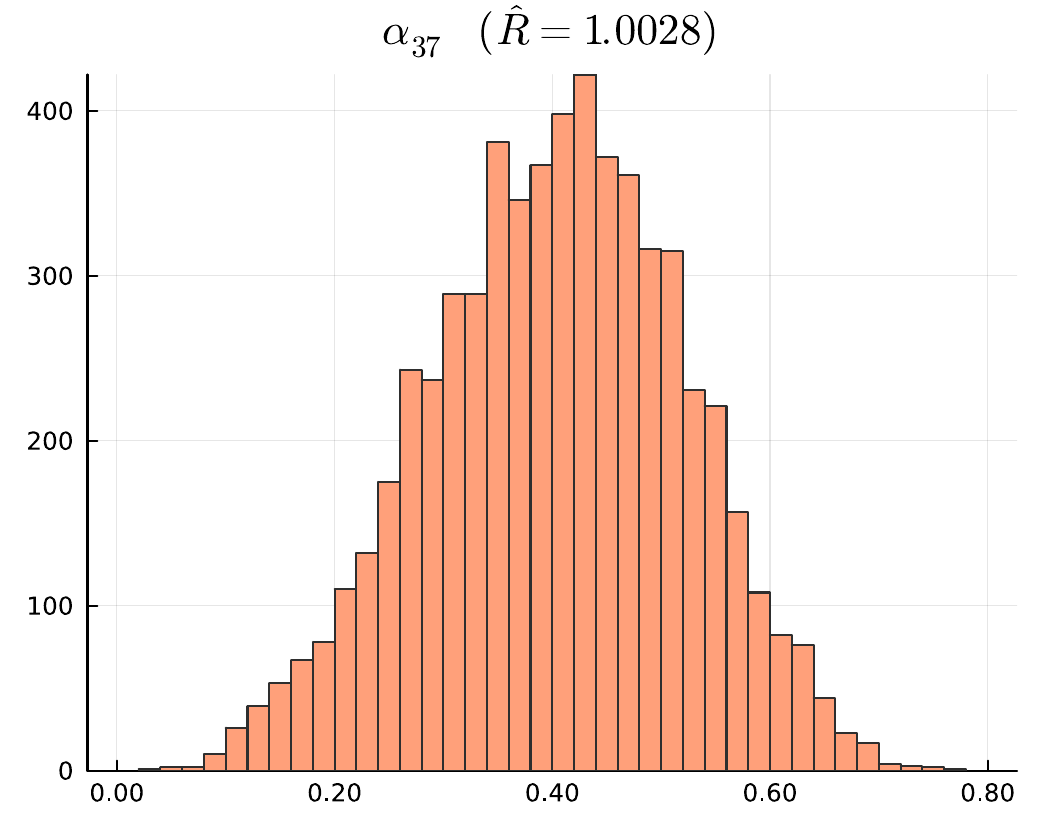}
    \caption{(Left) test log-likelihoods of GG and GA PHIBP models. (Middle) posterior distribution of $\alpha_0$. (Right) posterior distribution of $\alpha_{37}$ (corresponding to San Francisco).}
    \label{fig:mcmc_mix}
\end{figure}

Figure~\ref{fig:mcmc_mix} (left) compares the GG and GA models in terms of test log-likelihood, computed following \cite{hibp25}, and shows that GG clearly outperforms GA, likely due to its improved ability to capture rare diseases. Figure~\ref{fig:mcmc_mix} (middle) and (right) present the posterior distributions of $\alpha_0$ (global) and $\alpha_{37}$ (San Francisco), along with their $\hat{R}$ values~\cite{BrooksGeneral,GelmanInference}, where values close to 1 indicate good mixing. The posterior for $\alpha_0$ is concentrated near zero, suggesting limited diversity of rare diseases overall, whereas $\alpha_{37}$ is centered around approximately 0.4, indicating substantially greater rare-disease diversity within San Francisco.

\begin{figure}
    \centering
    \includegraphics[width=\linewidth]{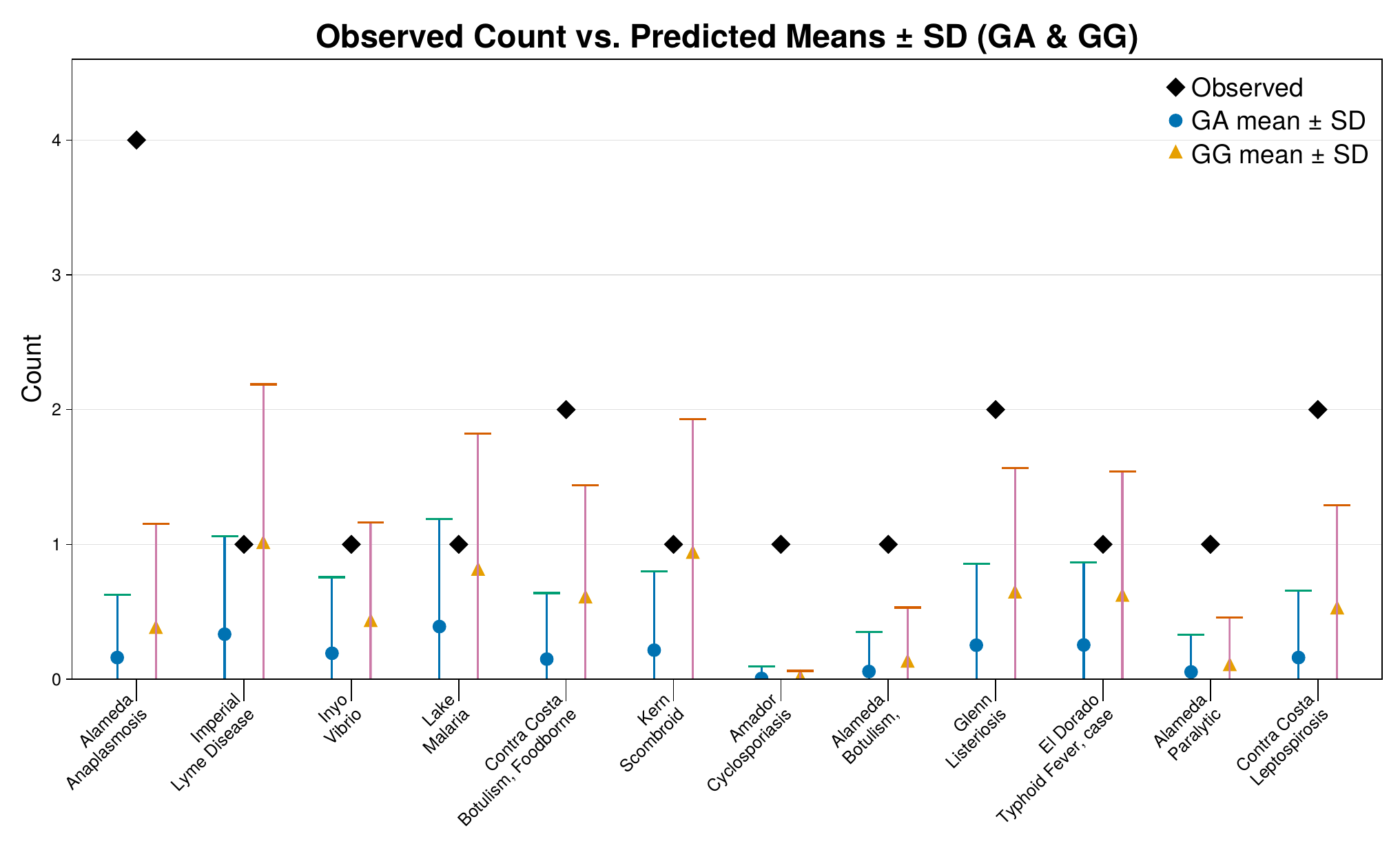}
    \caption{Predicted vs test data counts for county-disease pairs with 0 counts in training dataset.}
    \label{fig:pred_zero} 
\end{figure}

An appealing feature of the PHIBP model is its ability to predict novel diseases, i.e., to assign positive counts to diseases that were never observed for a given county in the training data. To illustrate this capability, and to contrast the behavior of GG and GA variants, we identify all (county, disease) pairs with zero training-set counts and examine their predicted counts in the test set. Because there are several hundred such pairs, we subsample 12 and plot the predicted counts from the GG and GA models against the true test-set counts in Figure~\ref{fig:pred_zero}. As shown, both models assign nonzero predictions to unseen diseases—an indication of information sharing across groups—but GG typically produces higher estimates that more closely match the observed counts, consistent with its diversity-promoting properties. Similar behavior was reported in the microbiome experiments of \cite{hibp25}.

% Figure \ref{fig:pred_zero} illustrates the predicted counts for county-disease pairs which have zero observed counts in the train data, but nonzero in the test data. We see that both GA and GG predict nonzero counts, which indicates a borrowing of information between groups. It appears as well that GG tends to predict higher counts than GA, {\color{blue} connect to original PHIBP paper}.

\begin{figure}
    \centering
    \includegraphics[width=\linewidth]{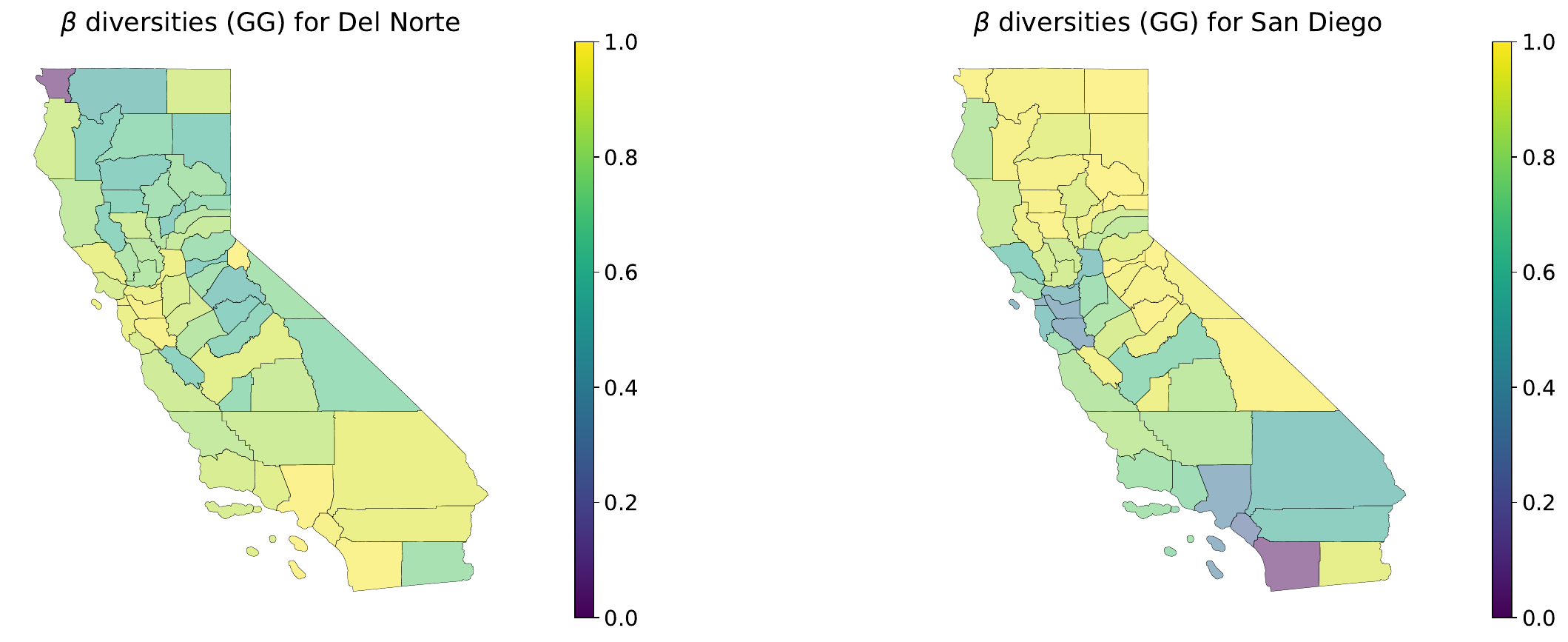}
    \caption{Heatmap of posterior mean of pairwise beta-diversities for GG with reference counties (in dark blue) as Del Norte (left) and San Diego (right).}
    \label{fig:beta_div}
\end{figure}

We now proceed to investigate posterior inference on the alpha/beta diversities as a function of geographical location. As beta-diversity is a measure of between-group diversity, we will investigate the beta-diversity for GG with respect to a reference county, which can be interpreted as a single row in a county by county beta-diversity matrix. Figure \ref{fig:beta_div} presents two geographical heatmaps of the counties in California, where the shade represents the posterior mean of the beta-diversities with respect to two reference counties, namely Del Norte (left) and San Diego (right).  The darker shade indicates a low beta-diversity, that is similarity in infectious disease profile, and the reference counties are in dark blue as they have 0 beta-diversity when compared to themselves. We visually see a clear trend: for counties near the reference counties, the beta-diversity tends to be lowest in the surrounding regions, with maximum beta-diversity as we move far away from the reference county. This is an intuitive finding, given that infectious diseases are transmissible between people, and geographic vicinity is likely a large driver in this transmission.  

\begin{figure}
    \centering
    \includegraphics[width=\linewidth]{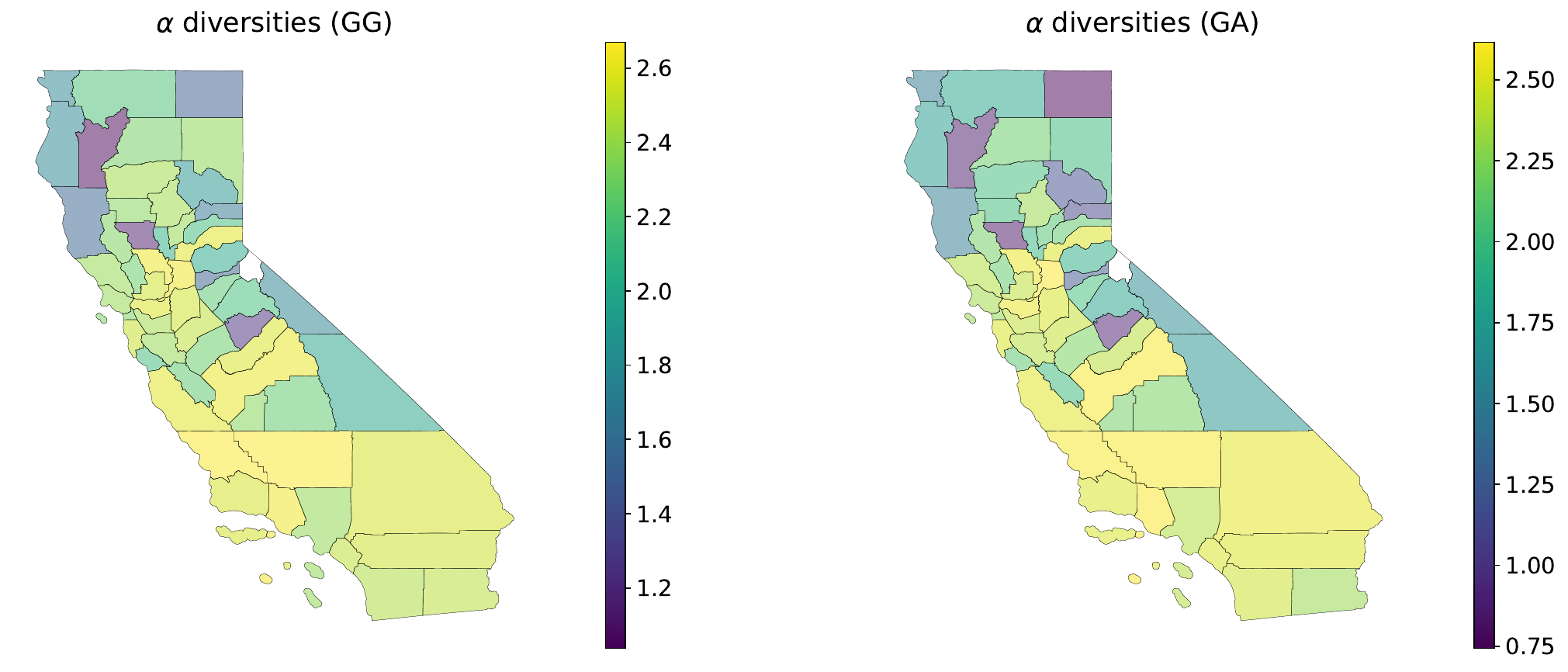}
    \caption{Heatmap of posterior mean of alpha-diversities for GG (left) and GA (right).}
    \label{fig:alpha_div}
\end{figure}

\begin{figure}
    \centering
    \includegraphics[width=\linewidth]{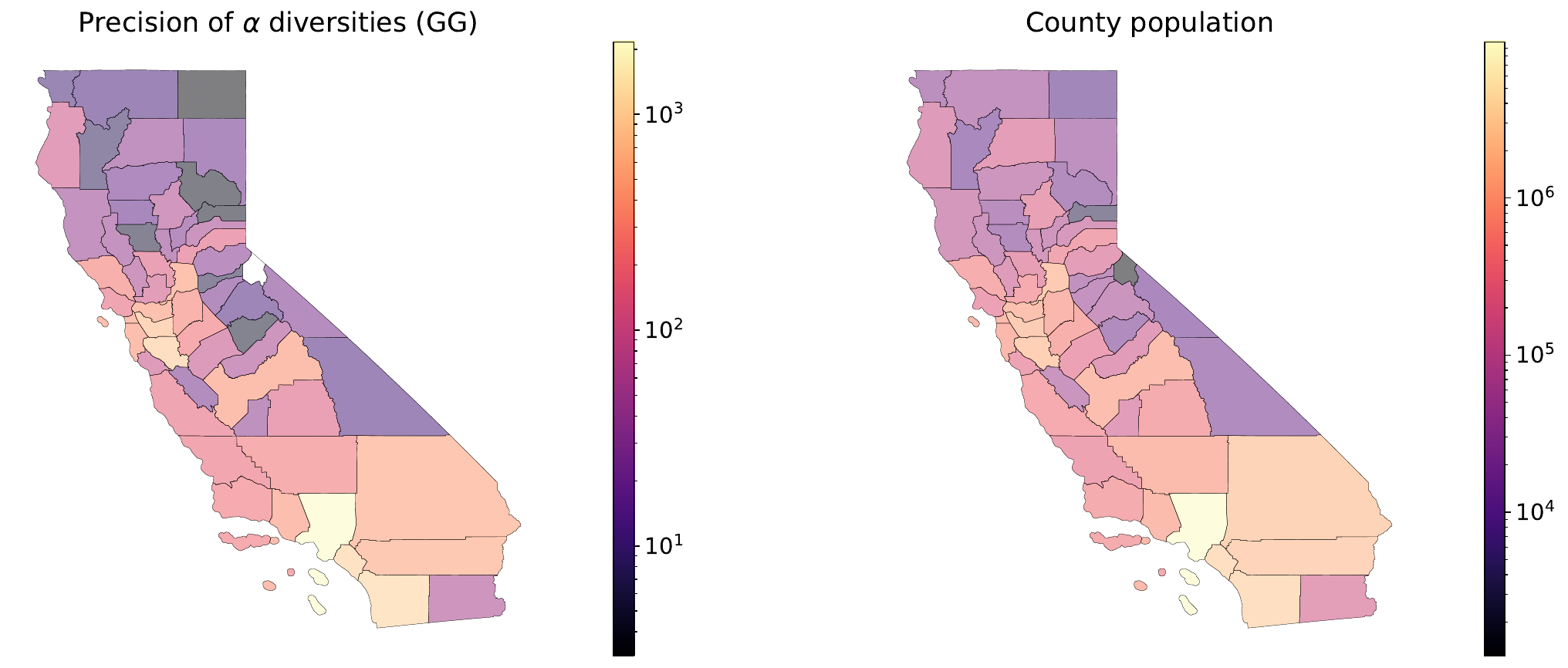}
    \caption{Heatmap of posterior precision of alpha-diversities for GG (left) and average county population from 2001-2014 (right); note the logarithmic scale.}
    \label{fig:alpha_prec}
\end{figure}

Figure \ref{fig:alpha_div} presents a similar geographical heatmap for alpha-diversities in each county, where we now compare GG and GA. We again see a geographical clustering in the alpha-diversity in Figure \ref{fig:alpha_div}, where southern counties tend to exhibit higher alpha-diversity indicating a richer disease profile. We also see in Figure \ref{fig:alpha_div} that GG tends to display slightly higher levels of alpha-diversity as expected, which is in agreement with the discussion in Section \ref{sec:exp_det} and the results of \cite{hibp25}. Finally, Figure \ref{fig:alpha_prec} (left) conveys the posterior uncertainty in alpha-diversities, namely the {precision} of the posterior distribution over the alpha-diversities. We see a direct connection of posterior uncertainty to the average county population between 2001-2014 plotted in Figure \ref{fig:alpha_prec} (right), which accurately depicts largest posterior precision for counties with the greatest population such as Los Angeles.


\section{Extensions: From Local Borrowing to Global Architectures}\label{sec:extensions}

The PHIBP framework demonstrated here for infectious disease prediction represents a specialized instance of a more fundamental mathematical architecture developed in~\cite{james2025coagfragduality}, where $Z$ represents a coarse process and $I$ a coupled fine process, both embedded within a four-component simultaneous structural duality system. While our focus has been on the practical challenge of predicting disease outbreaks from sparse data, the underlying construction connects to deeper principles of information sharing and structural duality---a concept introduced in that work---in hierarchical systems.

\subsection{The Architecture of Information Borrowing}

The success of PHIBP in predicting unseen diseases and capturing geographic clustering in diversity measures reflects a specific realization of what is termed ``cloud duality'' in~\cite{james2025coagfragduality}---the principle that hierarchical borrowing of information corresponds to reversible coagulation-fragmentation operations on the underlying point processes. 

In our disease application, this manifests as the model's ability to:
\begin{itemize}
\item Fragment the global disease pool into county-specific patterns (via the compound Poisson decomposition in equation~\eqref{eq:joint-law})
\item Coagulate local observations to inform global patterns (through the hierarchical random measure structure)
\end{itemize}

This bidirectional flow---visible in how San Francisco's higher diversity parameter $\alpha_{37} \approx 0.4$ versus the global $\alpha_0 \approx 0$ influences predictions---is not merely statistical borrowing but a structural property of the underlying mixed Poisson architecture that enabled our model to assign nonzero predictions to county-disease pairs with zero training counts.

\subsection{Practical Extensions for Disease Surveillance}

The computational efficiency demonstrated in our California analysis, where both GG and GA models achieved stable convergence, suggests several immediate extensions:

\textbf{Multi-resolution modeling}: While we analyzed county-level data across 58 counties, the framework naturally extends to hierarchical structures like State $\to$ County $\to$ ZIP code, with each level maintaining its own subordinator process. The mixed truncated Poisson structure ensures tractability even as complexity grows.

\textbf{Dynamic surveillance}: Although our analysis treated years 2001--2014 independently for training, the framework's L\'evy-It\^o foundation~\cite{james2025coagfragduality} enables continuous-time modeling where disease emergence follows time-evolving Poisson random measures, capturing seasonal patterns and emerging variants.

\textbf{Alternative disease spaces}: The beta-diversity patterns we observed arose from geographic proximity, but the framework can incorporate other distance metrics---genomic similarity between strains, human mobility networks, or environmental factors---by choosing appropriate L\'evy measures $\tau_j$ that encode domain-specific transmission dynamics~\cite[Section 7.5.3]{james2025coagfragduality}.

\subsection{From Local Patterns to Global Understanding}

Our analysis revealed clear geographic structure in disease diversity: southern counties exhibited higher alpha-diversity, while beta-diversity increased with geographic distance from reference counties. The precision of these estimates scaled with population size, demonstrating how the model appropriately weights information from data-rich versus data-sparse regions.

These patterns emerge from the four-component system $(I_j, A_j, F_{j,\ell}, Z_j)_{j \in [J]}$ where:
\begin{itemize}
\item The coarse process $Z$ captures the observed disease counts (our 36 rare diseases)
\item The fine process $I$ represents unobserved transmission chains
\item The allocation process $A$ determines which diseases manifest in each county
\item The fragmentation operators $F_{j,\ell}$ describe how statewide patterns decompose locally
\end{itemize}

This architecture enabled the key finding that even county-disease pairs with zero training counts could be predicted with meaningful uncertainty quantification---a capability essential for operational surveillance where the question ``where will it appear next?'' is paramount.

\subsection{Implications for Real-Time Implementation}

The framework's robustness with sparse data---demonstrated by accurate predictions for diseases never observed in training---combined with the exact sampling procedures, enables:
\begin{itemize}
\item \textbf{Nowcasting}: Real-time estimation of current outbreak intensity from incomplete reporting
\item \textbf{Forecasting}: Principled prediction to currently unaffected counties via the allocation machinery  
\item \textbf{Uncertainty quantification}: Full posterior distributions for resource allocation decisions
\end{itemize}

The same architecture that allows borrowing of statistical strength across California's 58 counties could be applied to multi-site clinical trials, environmental monitoring networks, or financial contagion modeling---each instantiating the same fundamental duality with domain-appropriate specifications~\cite[Section 7.5]{james2025coagfragduality}.

These capabilities, grounded in the deeper mathematical architecture of simultaneous structural duality, point toward a unified framework for understanding how epidemiological information propagates through hierarchical systems---whether predicting the next county to report Listeriosis or identifying emerging transmission patterns---where the challenge is not just handling zeros, but understanding what they represent.

\section{Conclusion}

This work demonstrated the Poisson Hierarchical Indian Buffet Process (PHIBP) framework's effectiveness in modeling rare infectious diseases across California's 58 counties from 2001-2019. By analyzing 36 diseases with extreme sparsity of county-disease-year combinations, we showed how principled hierarchical modeling can extract meaningful patterns where traditional methods fail.

Our experimental results compared two specific PHIBP variants from the broader family of possible specifications: the gamma-gamma (GG) and gamma-alpha (GA) models. Among these two choices, GG consistently outperformed GA in test log-likelihood, demonstrating superior ability to capture rare disease patterns. Both models successfully predicted disease occurrences for county-disease pairs never observed in training data—a critical capability for surveillance systems where anticipating disease emergence in new locations is paramount. These GG and GA specifications represent only a small subset of the rich family of Lévy measures available within the PHIBP framework. The h-biased framework presented in Section 7 of \cite{james2025coagfragduality} offers even greater flexibility, allowing for asymmetric allocation patterns and size-biased sampling that could better capture disease-specific transmission characteristics or preferential attachment dynamics in epidemic spread.

Geographic structure emerged clearly in our diversity analyses: beta-diversity increased with geographic distance between counties, while alpha-diversity showed regional clustering with southern counties exhibiting richer disease profiles. The precision of these estimates scaled appropriately with population size, with densely populated counties like Los Angeles showing the highest posterior certainty. These patterns, while robust under both GG and GA specifications, illustrate how different choices of subordinator processes can emphasize different aspects of the hierarchical structure.

These empirical successes reflect the model's dual capacity to fragment global disease pools into county-specific patterns while simultaneously borrowing strength across locations to inform predictions. San Francisco's elevated diversity parameter ($\alpha_{37} \approx 0.4$) compared to the global baseline ($\alpha_0 \approx 0$) exemplifies how the framework captures local heterogeneity while maintaining computational tractability through the mixed truncated Poisson representation.

The PHIBP framework's success on this challenging infectious disease dataset demonstrates that its underlying mathematical architecture—grounded in coagulation-fragmentation duality—provides a principled approach to information sharing in hierarchical systems. The h-biased extensions could prove particularly valuable for modeling superspreader events or hub counties that disproportionately influence disease transmission networks. Future applications might explore multi-resolution disease surveillance, continuous-time outbreak modeling, and incorporation of alternative distance metrics based on genomic similarity or mobility networks, potentially leveraging the h-biased framework's ability to model preferential allocation patterns. More broadly, this unified framework for understanding how information propagates through complex hierarchical systems extends beyond epidemiology to microbial communities, clinical trials, and other structured populations where the challenge is not just handling zeros, but understanding what they mean.

%%%%%%%%%%%%%%%%%%%%%%%%%%%%%%%%%%%%%%%%%%%%%%
%% Single Appendix:                         %%
%%%%%%%%%%%%%%%%%%%%%%%%%%%%%%%%%%%%%%%%%%%%%%
%\begin{appendix}
%\section*{???}%% if no title is needed, leave empty \section*{}.
%\end{appendix}
%%%%%%%%%%%%%%%%%%%%%%%%%%%%%%%%%%%%%%%%%%%%%%
%% Multiple Appendixes:                     %%
%%%%%%%%%%%%%%%%%%%%%%%%%%%%%%%%%%%%%%%%%%%%%%
%\begin{appendix}
%\section{???}
%
%\section{???}
%
%\end{appendix}

%%%%%%%%%%%%%%%%%%%%%%%%%%%%%%%%%%%%%%%%%%%%%%
%% Support information, if any,             %%
%% should be provided in the                %%
%% Acknowledgements section.                %%
%%%%%%%%%%%%%%%%%%%%%%%%%%%%%%%%%%%%%%%%%%%%%%
%\begin{acks}[Acknowledgments]
% The authors would like to thank ...
%\end{acks}
%%%%%%%%%%%%%%%%%%%%%%%%%%%%%%%%%%%%%%%%%%%%%%
%% Funding information, if any,             %%
%% should be provided in the                %%
%% funding section.                         %%
%%%%%%%%%%%%%%%%%%%%%%%%%%%%%%%%%%%%%%%%%%%%%%
\begin{funding}
This work was supported in part by RGC-ECS grant 27304424 and RGC-GRF grants 16301521, 17306925 of the HKSAR.
\end{funding}

%%%%%%%%%%%%%%%%%%%%%%%%%%%%%%%%%%%%%%%%%%%%%%
%% Supplementary Material, including data   %%
%% sets and code, should be provided in     %%
%% {supplement} environment with title      %%
%% and short description. It cannot be      %%
%% available exclusively as external link.  %%
%% All Supplementary Material must be       %%
%% available to the reader on Project       %%
%% Euclid with the published article.       %%
%%%%%%%%%%%%%%%%%%%%%%%%%%%%%%%%%%%%%%%%%%%%%%
%\begin{supplement}
%\stitle{???}
%\sdescription{???.}
%\end{supplement}

%%%%%%%%%%%%%%%%%%%%%%%%%%%%%%%%%%%%%%%%%%%%%%%%%%%%%%%%%%%%%
%%                  The Bibliography                       %%
%%%%%%%%%%%%%%%%%%%%%%%%%%%%%%%%%%%%%%%%%%%%%%%%%%%%%%%%%%%%%

%% if your bibliography is in bibtex format, uncomment commands:
%\bibliographystyle{imsart-number} % Style BST file (imsart-number.bst or imsart-nameyear.bst)
%\bibliography{bibliography}       % Bibliography file (usually '*.bib')

%% or include bibliography directly:
% \begin{thebibliography}{}
% \bibitem{b1}
% \end{thebibliography}

%%%%%%%%%%%%%%%%%%%%%%%%%%%%%%%%%

%\bibliographystyle{agsm}
%\bibliography{bibliography}
\bibliographystyle{imsart-number} 
%\bibliography{bibliography2}   
\bibliography{bibliographyfullupdated}
\end{document}